\pdfoutput=1

\documentclass[11pt]{article}

\usepackage{tcolorbox}
\tcbuselibrary{listings,skins}
\usepackage{lipsum} 

\tcbset{
    promptbox/.style={
        colback=white,
        colframe=black,
        fonttitle=\bfseries,
        title={Prompt for Notes Writing},
        arc=0pt,
        outer arc=0pt,
        boxrule=1pt,
        left=6pt,
        right=6pt,
        top=6pt,
        bottom=6pt,
        listing only,
        listing options={
            basicstyle=\ttfamily,
            breaklines=true,
            showstringspaces=false,
            columns=fullflexible,
            escapeinside={(*@}{@*)}
        }
    }
}

\usepackage[final]{acl}

\usepackage{times}
\usepackage{latexsym}

\usepackage[T1]{fontenc}

\usepackage[utf8]{inputenc}

\usepackage{microtype}

\usepackage{inconsolata}

\usepackage{graphicx}

\usepackage{threeparttable} 

\usepackage{xurl}

\title{Model Fusion with Multi-LoRA Inference for Tool-Enhanced Game Dialogue Agents}

\author{Kangxu Wang, Ze Chen,  Chengcheng Wei, Jiewen Zheng, \\ \bf{Jiarong He}, \bf{Max Gao}\\
        Interactive Entertainment Group of Netease Inc., Guangzhou, China\\
        \{wangkangxu, jackchen, weichengcheng, zhengjiewen, gzhejiarong, jgao\}@corp.netease.com}

\begin{document}
\maketitle
\begin{abstract}
This paper presents the \textbf{opdainlp} team's solution for 
the GPU track of the CPDC 2025 challenge. The challenge consists of three tasks, aiming to build an in-game conversational AI that adheres to character personas, aligns with the game's worldview, and supports function calling.
Considering both effectiveness and resource/time constraints during inference, we synthesized data for some of the tasks based on the datasets provided by the competition organizers. We employed Qwen3-14B with LoRA fine-tuning and model fusion, and utilized a base model integrated with multiple LoRA adapters during inference.
Specifically, in the competition, we used three distinct LoRA adapters to handle tool calling, response generation with tool call results, and response generation without tool call results, respectively. MultiLoRA inference was implemented using vLLM.
Our solution achieved the first place in Task 1 and Task 3, and the second place in Task 2 of the GPU track.
\end{abstract}

\section{Introduction}
Since the release of ChatGPT, large language models (LLMs) have been widely applied across various fields due to their strong performance in a wide range of tasks compared to traditional models.
Since large language models (LLMs) are typically trained on massive corpora during pretraining—including extensive novels, scripts, and other texts containing dialogues of various characters—and are further trained to follow instructions during post-training, they exhibit strong capabilities in role-playing. This enables users to experience conversational systems that feel significantly more human-like compared to traditional models.
%

%
CPDC 2025, hosted by Sony,\footnote{\url{https://www.aicrowd.com/challenges/commonsense-persona-grounded-dialogue-challenge-2025}} focuses on evaluating dialogue interaction systems developed by participants for use in player-NPC interactions within games \cite{gao-etal-2023-peacok}.
Compared to general human-machine conversational chat scenarios, in gaming contexts, players place greater emphasis on the integration of NPCs within the overall game environment—specifically, they expect not only casual conversations with NPCs, but also interactions involving task-related and knowledge-based elements.
%

The competition has set up three sub-tasks:
Task 1: Task-Oriented Dialogue Agents, Task 2: Context-Aware Dialogue Agents, and Task 3: Integrating Contextual Dialogue and Task Execution.
Each task has two tracks: GPU and API. In the GPU track, participants can upload fine-tuned models for inference during the testing phase, while the API track only allows predictions using the OpenAI GPT-4o-mini interface. 
In this competition, we primarily focused on 
the GPU track. For the GPU track, each submission runs on an AWS g6e.2xlarge node with 8 vCPUs, 64 GB RAM, and one NVIDIA L40s GPU with 48 GB GPU memory. The timeout for each turn is set at 7 seconds.
The key difference between Task 1 and Task 2 is that Task 1 requires function calling to support game NPC features like item queries and sales, while Task 2 focuses on character interaction based on persona and background knowledge without function calling. Task 3 evaluates both Task 1 and Task 2 test sets and takes their average for the final result. The organizers provided 40 conversations each for Task 1 and Task 2 as training data.

To address the complexity of game-domain dialogue, \citet{chen-etal-2025-tcqa2} proposes a collaborative multi-agent architecture, TCQA$^2$, in which specialized agents handle distinct sub-tasks, enabling high precision, low latency, personalized interaction, and robust safety. Inspired by this approach, in this competition, we employ different agents (models) depending on whether tool calls are required, allowing us to effectively address diverse scenarios.
In the following sections, we'll share some effective technical solutions we tried in this competition, including:

\begin{enumerate}
  \item Performing fine-tuning on separate datasets for function calling and different dialogue scenarios, and using MultiLoRA to support scenario-specific predictions.
   We adopted this approach because we found that merging datasets from different scenarios (e.g., tool calling and dialogue) affected performance in this low-sample setting. Additionally, resource constraints prevented deploying multiple full-parameter models. MultiLoRA allowed us to optimize intermediate processes for each task independently using scenario-specific datasets while meeting competition inference constraints.

  \item Model fusion: 
  Performance improvement through large language model fusion has been validated in multiple scenarios \cite{deotte2024winningamazonkddcup24,kim2024solar107bscalinglarge}. In this competition, we also adopted a model fusion approach and achieved noticeable performance gains.

  \item Data synthesis: For dialogue data, we used the original user inputs and generated assistant outputs using commercial large model APIs like GPT-4.1, Claude-Sonnet-4, and Qwen-Max to create new training data for model fine-tuning and fusion.

  \item We submitted and compared 
  performance of several non-fine-tuned models,
  ultimately selecting Qwen3-14B for fine-tuning.
\end{enumerate}

\section{Task Description}
In this section, we provide a detailed introduction to the overall input and output format of the three tracks in CPDC 2025. 
Although the evaluation data content differs across the three tasks, their data formats are identical. Specifically, for each dialogue turn, the input from the test set includes background information, dialogue context, and function-related information. During each prediction, the system may utilize any of the provided input information. 
The dialogue system is first required to predict a function call, then generates a response for the in-game NPC based on the function call result and the player's dialogue inputs.

\subsection{Input and Output Format}
The background information for each input consists of five parts: worldview, persona, role, knowledge, and state, which remain fixed throughout a given conversation. 
The ``worldview'' is a long text describing the game's worldview information, including foundational rules such as currency units and task difficulty levels. The ``persona'' contains basic information about the NPC, such as name, age, gender, occupation, and appearance. The ``role'' defines the NPC's character role and functional purpose within the game; in the provided training set, all NPCs are weapon shop merchants. The ``knowledge'' component includes game-related information relevant to the conversation, containing two parts: ``general\_info'', which covers detailed game settings related to the dialogue—such as ``Guild and Environment'' and ``Weapons and Maintenance''—explaining the relationship between players (adventurers in the game) and weapons, and justifying why players would interact with a weapon merchant NPC; and ``knowledge\_info'', which contains information about the weapons available in the shop, including weapon name, type, and description, with approximately 20 weapons included per conversation. The ``state'' field specifies the current location, time, and weather conditions under which the dialogue takes place.

For each conversation in the dataset, a \texttt{function\_list\_id} is provided. This identifier allows participants to retrieve a candidate subset of functions from the official function registry released by the organizers. 
This subset is intended to guide models in performing tool-augmented response generation. The functions within each subset are categorized into two types: 
\texttt{action\_functions} and \texttt{tool\_functions}.
\texttt{action\_functions} correspond to in-game actions that an NPC may perform, such as \texttt{sell} or \texttt{select the specified quest}. 
In contrast, \texttt{tool\_functions} are used to retrieve information like item attributes or prices. 
In Task 1 and Task 3, models are expected to predict
 functions 
  during dialogue generation. 
  However, in Task 2, although the \texttt{function\_list\_id} is still provided in the data, no dialogues in the training set require function prediction.

\section{System Overview}

\begin{figure*}
    \centering
    \includegraphics[width=1\linewidth]{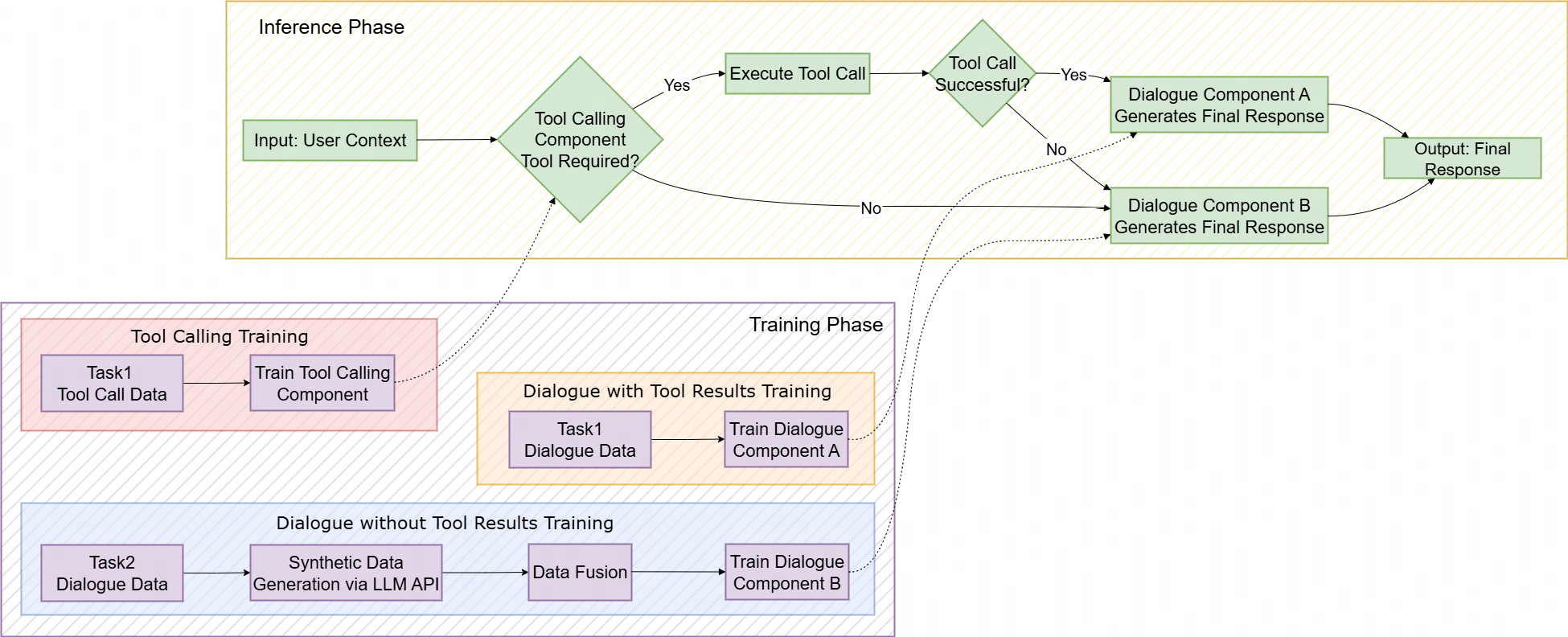}
    \caption{High-Level Oerview of our pipeline for the CPDC 2025}
    \label{fig:pipeline}
\end{figure*}

In the early stage of the competition, we evaluated various open-source models on Task 1 using the baseline prediction code. Considering both strong performance on Task 1 and reliable behavior on Task 2, we ultimately adopted Qwen3-14B as the base model for fine-tuning across all three tasks.

We divide the overall task into three sub-scenes: (1) tool-calling scenarios, (2) dialogue scenarios with tool-calling results, and (3) dialogue scenarios without tool-calling results.

For the tool-calling scenario, we constructed the dataset using only the tool-calling prediction portion from Task 1's training set, which does not involve dialogue prediction.
%
For dialogue scenarios with tool call results, the dialogue portion of the data from Task 1 was used for training.
For dialogue scenarios without tool calling, 
we trained using 
Task 2's provided dialogue data 
and 
synthetic data using commercial LLM APIs
separately, 
and performed 
model fusion.

During actual inference, we classify a scenario as involving tool calling if the tool-calling component yields any results.
As defined by the competition, the Task 3 score is computed by averaging predictions made on the Task 1 and Task 2 test sets via submission to the Task 3 system. Given that our best results on both Task 1 and Task 2 were obtained through this same system, we focus exclusively on describing the design of our Task 3 system in this paper.

For training in the tool-calling scenario, we constructed the training data following the tool-calling format of Hermes.\footnote{\url{https://github.com/NousResearch/Hermes-Function-Calling}}
Specifically, we included the candidate function information in the \texttt{system\_prompt} using the Hermes format, and included the dialogue history between the user and the NPC, the current input, and weapon-related knowledge from the dialogue context in the \texttt{user\_input}. 
Our experiments showed that including redundant information such as game worldview or world background degraded model performance, while incorporating dialogue history improved the model's ability to identify functions when coreference cues were missing in the current turn. Including weapon information enhanced the model's capability in extracting function arguments. We used only the dataset provided by the competition organizers for training. After training, we performed parameter averaging over the LoRA checkpoints from each epoch, which further improved performance in our experiments.
%

In our experiments, the main difference between handling dialogue scenarios with tool calling recognition and those without tool calling lies in the information being transmitted.
For scenarios involving tool recognition, our experimental observations showed that dialogues tend to be knowledge-oriented. When constructing the training set model inputs, we included essential elements such as role, persona, and function call results, along with weapon/item information (weapon descriptions help NPCs provide targeted responses), while excluding game worldview and lore.
For dialogue scenarios without tool calling, we excluded weapon/item information and instead only included game worldview and lore. 
Furthermore,
for scenarios with tool-calling recognition, we trained exclusively using the data provided by the competition organizers; for scenarios without tool calling, we supplemented the official data by synthesizing additional data using GPT-4.1, Claude-Sonnet-4, and Qwen-Max. 
For both modules, we used LoRA parameter fusion to merge the models.

For data synthesis in the dialogue scenario without tool calls, we fed the input from the original training data of Task 2 into commercial LLM APIs to generate responses. 
We tested two strategies for response generation: one in which the entire dialogue history is provided at once for response generation, and another in which the turns from the original training data are fed sequentially, with the dialogue history updated by the commercial LLM's generated response at each step. Our evaluation showed that the second synthesis approach yielded better performance when used for training. 

We describe the details of the prompts used in the three scenarios in Appendix \ref{sec:appendix}.

\section{Experiments}
\subsection{Experimental Setup}
We conducted all training and offline evaluation on 8 NVIDIA A30 GPUs. During the selection of the fine-tuning base model, we used the provided baseline code to evaluate several open-source models on Task 1, including \texttt{Llama-3.1-8B-Instruct}~\cite{grattafiori2024llama3herdmodels}, \texttt{gemma-3-12b-it}~\cite{gemmateam2025gemma3technicalreport}, \texttt{Qwen2.5-14B-Instruct}~\cite{yang2024qwen2technicalreport}, \texttt{Qwen3-14B}, \texttt{Qwen3-30B-A3B}~\cite{yang2025qwen3technicalreport} and \texttt{Llama-xLAM-2-8b-fc-r}~\cite{zhang2024xlamfamilylargeaction}. 
We also evaluated the performance of quantized Qwen2.5 series 32B models. Due to hardware limitations that prevented frequent fine-tuning of 32B models, we did not continue evaluating quantized models from the Qwen3 series after their release.

\begin{table}
  \centering
  \begin{tabular}{ll}
    \hline
    \textbf{Model}           & \textbf{Score} \\
    \hline
    Llama-3.1-8B-Instruct      & 0.329           \\
    gemma-3-12b-it     & 0.384        \\
    Qwen2.5-14B-Instruct       & 0.449           \\
    Qwen3-14B & 0.457 \\
        Qwen2.5-32B-Instruct-GPTQ-Int4 & 0.433 \\
        Qwen2.5-32B-Instruct-AWQ-int4 & 0.436 \\
        Qwen2.5-32B-Instruct-GPTQ-Int8 & 0.425 \\
        Llama-xLAM-2-8b-fc-r & 0.463 \\
        Qwen3-30B-A3B & \bf{0.498} \\
        Qwen3-30B-A3B (no think) & 0.491 \\
    \hline
  \end{tabular}
  \caption{\label{table1}
    Performance of different models on the baseline in Task 1 (Round 1).
  }
\end{table}

In this competition, adjusting the model's input information had a significant impact on the overall results. Before fine-tuning, we discovered that the initial baseline didn't include dialogue history during function calling. After adding 
historical context, we achieved improved performance.

\begin{table}[h]
  \centering
  \begin{tabular}{lccc}
    \hline
         &  Task3 &  Task1 &  Task2 \\
    \hline
        w/o history & 0.526 & 0.457 & \bf{0.595} \\
        w/ history & \bf{0.536} & \bf{0.478} & \bf{0.595} \\
    \hline
  \end{tabular}
  \caption{\label{table2}
    Scores with and without dialogue history during function calling (Round 1).
  }
\end{table}

During the fine-tuning phase, 
our training configuration remained consistent across all scenarios. We employed LoRA for fine-tuning, with both LoRA rank and LoRA alpha set to 128. The weight decay was configured to 0.1, the warmup ratio to 0.05, and the type of learning rate scheduler was specified as cosine.
We adopted the combination of DeepSpeed ZeRO-3 and FlashAttention, enabling the fine-tuning of the Qwen3-14B model on A30 GPUs. During training, we set the batch size to 8 and trained for 3 epochs. At the end of each epoch, a LoRA checkpoint was saved. 
After completing the training, we averaged the weights of these three LoRA checkpoints to obtain the final checkpoint. Our entire training process was implemented using ms-swift \cite{zhao2025swiftascalablelightweightinfrastructure}.

\begin{table}[h]
  \centering
  \begin{tabular}{lccc}
    \hline
       Strategy  &  Task3 &  Task1 &  Task2 \\
    \hline
    No Fine-tuning & 0.536 & 0.478 & \bf{0.595} \\
        LoRA 1 Epoch & 0.554 & 0.518 & 0.592 \\
        LoRA 2 Epochs & 0.556 & 0.522 & 0.591 \\
        LoRA 3 Epochs & 0.546 & 0.508 & 0.586 \\
        LoRA Fusion & \bf{0.562} & \bf{0.536} & 0.588 \\
    \hline
  \end{tabular}
  \caption{\label{table3}
  Scores of LoRA Fine-tuning and Fusion Strategies on Function Call (Round 1).
  }
\end{table}

We continuously adjusted the composition of the model's input for fine-tuning, and discovered that providing background knowledge about weapons during the function-calling stage significantly enhanced performance.

\begin{table}[h]
  \centering
  \begin{tabular}{cccc}
    \hline
      Weapon Information   &  Task3 &  Task1 &  Task2 \\
    \hline
    w/o Weapon Info & 0.562 & 0.536 & \bf{0.588} \\
        w/ Weapon Info      & \bf{0.635} & \bf{0.682} & \bf{0.588} \\
    \hline
  \end{tabular}
  \caption{\label{table4}
    Effect of including weapon information during Function Call on Final Scores.
  }
\end{table}

In Round 1, we attempted to fine-tune a single LoRA adapter for NPC responses, shared across both cases with and without function call results. 
However, this approach performed worse than the
non-fine-tuned Qwen3-14b model. 
Through testing, we discovered that the automatic evaluation result for Task 1 showed high correlation with the training data, whereas Task 2 (which does not involve function calls) shows a lower correlation.
Based on these findings, in Round 2 we decoupled the modeling by training three separate LoRA adapters, two of which are dedicated to NPC response generation.
For the scenario without function calls, we further augmented the training data using responses generated via commercial model APIs.

We used commercial model APIs to regenerate model outputs based on the Task 2 training set. During generation, temperature was set to 0.1 and top\_p to 0.95. For each commercial API, we trained a separate LoRA adapter for 3 epochs and averaged the checkpoints across epochs. Finally, we averaged the resulting LoRA weights from all APIs to obtain the final combined adapter.

\begin{table}[h]
  \centering
  \begin{tabular}{l c}
    \hline
      Training Data & Score \\
    \hline
Original Task 2 Training Set & 0.587 \\
GPT-4.1-2025-04-14 SD & 0.607 \\
Claude-Sonnet-4-20250514 SD & 0.608 \\
Qwen-Max SD & 0.611 \\
\textbf{Average LoRA Weights} & \bf{0.615} \\
    \hline
  \end{tabular}
  \caption{\label{table5}
    Effect of Synthetic Data (SD) using different models on Task 2 Performance.
  }
\end{table}

\begin{table}[h]
  \centering
  \begin{tabular}{lccc}
    \hline
         &  Task3 &  Task1 &  Task2 \\
    \hline
      Final Results & 0.635 & 0.655 & 0.615 \\
    \hline
  \end{tabular}
  \caption{\label{table6}
    Automatic Evaluation Results of Task 3 on Round 2 Leaderboard After Model Combination.
  }
\end{table}

  
  
  
  

\subsection{Evaluation}
In CPDC2025, the evaluation of Task 1 directly depends on an automatic score which is the average score of Function Score and BLEURT Score~\cite{sellam-etal-2020-bleurt}. 
%
The evaluation of Task 2
and Task 3
is divided into two stages, determined by automatic scoring and human evaluation rankings. 
For Task 2, the automatic score is the average of 
CPDC Score \cite{wakaki2024comperdialcommonsensepersonagroundeddialogue} and BLEURT Score;
and human evaluation is based on Response Quality and Knowledge Consistency.
For Task 3, 
the automatic score is calculated as the average of the automatic scores from Task 1 and Task 2. The final score in human evaluation is determined by the sum of ranks from each task.

We trained the model on the Task 2 dataset and used it for generating responses without function calls in Task 3. 
To generate responses with function calls, we trained using only the original Task 1 training set.

\subsection{Results}

\begin{table}[h]
  \centering
  \begin{tabular}{lc}
    \hline
        Team &  Automatic Score \\
    \hline
      \textbf{opdainlp} & \bf{0.640} \\
      test\_team & 0.640 \\
      zvers & 0.632 \\
    \hline
  \end{tabular}
  \caption{\label{table7}
     Final Evaluation Results of the Task 1 GPU Track.
  }
\end{table}

\begin{table*}[h]
  \centering
  \begin{tabular}{l c c c c}
    \hline
        Team & Automatic Score & Sum of Rank & Response Rank & Knowledge Rank \\
    \hline
budai & 0.618 & 3 & 1 & 2 \\
\textbf{opdainlp} & 0.619 & 5 & 4 & 1 \\
test\_team & 0.597 & 8 & 3 & 5 \\
    \hline
  \end{tabular}
  \caption{\label{table8}
     Final Evaluation Results of the Task 2 GPU Track.
  }
\end{table*}


\begin{table*}[h]
  \centering
  \begin{tabular}{l c c c c}
    \hline
Team & Automatic Score & Sum of Rank & Task 1 Auto Rank & Task 2 Human Rank \\
    \hline
\textbf{opdainlp} & \bf{0.628} & 4 & 1 & 3 \\
zvers & 0.626 & 5 & 3 & 2 \\
MSRA\_SC & 0.600 & 6 & 5 & 1 \\
    \hline
  \end{tabular}
  \caption{\label{table9}
     Final Evaluation Results of the Task 3 GPU Track.
  }
\end{table*}

Table~\ref{table1} presents the performance of open-source models on Task 1. 
Qwen3-30B-A3B and the tool-use–optimized Llama-xLAM-2-8b-fc-r from Salesforce AI Research~\cite{zhang2024xlamfamilylargeaction} rank among the top performers on Task 1.
However, in our setup—where a single model is shared across tasks—Qwen3-30B-A3B suffers from response generation timeouts on Task 2 due to its large size. Similarly, Llama-xLAM-2-8b-fc-r exhibits significantly degraded performance on Task 2.
Among the remaining models, Qwen3-14B achieves the best performance, 
slightly outperforming
the 32B quantized variants.

Tables~\ref{table2} and~\ref{table7} demonstrate the impact of input formulation on overall performance. Incorporating dialogue history and knowledge information leads to substantial improvements in function call accuracy.
Tables~\ref{table3} and~\ref{table5}
present results from our data synthesis and model fusion approaches. With the limited training data available, both methods demonstrated significant improvements in overall system performance.

\section{Conclusion}
In this paper, we present our approach for CPDC-2025. 
We propose a MultiLoRA framework that integrates distinct fine-tuned adapters at different stages of the system. To address the challenge of limited training data, we employ techniques such as LoRA checkpoint averaging and synthetic data generation, which significantly improve the model's generalization capability. Our approach achieved first place in Tasks 1 and 3 on the GPU track, and second place in Task 2.


\bibliography{custom}

\clearpage
\newpage

\appendix

\section{Appendix}
\label{sec:appendix}

\subsection{Prompt Used in the Function Call Stage}
Only weapon information is provided, excluding unrelated knowledge such as worldview and character settings.

    {\centering
    \begin{tcolorbox}[title=System Prompt]
      You are a helpful assistant.\\
\\
\# Tools\\
\\
You may call one or more functions to assist with the state, conversation history, additional information, knowledge and user query.\\
\\
You are provided with function signatures within <tools></tools> XML tags:\\
<tools>\\
\{function\}\\
</tools>\\
\\
For each function call, return a json object with function name and arguments within <tool\_call></tool\_call> XML tags:\\
<tool\_call>\\
\{\{"name": <function-name>, "parameters": <args-json-object>\}\}\\
</tool\_call>\\
    \end{tcolorbox}}

    {\centering
    \begin{tcolorbox}[title=User Prompt]
      state:\\
\{state\}\\
\\
knowledge:\\
\{knowledge\_info\}\\
\\
anadditional information:\\
\{anadditional information\}\\
\\
conversation history:\\
\{history\}\\
\\
user query:\\
\{query\}
    \end{tcolorbox}}

\subsection{Prompt for Generating Responses with Function Call Results}
Character settings and knowledge are provided.

    {\centering
    \begin{tcolorbox}[title=System Prompt]
      \# Instruction\\
You are an assistant that plays the role of a character in a video game. \\
Use the following role-playing requirements, character settings and knowledge to create your response.\\
Try to keep your response to no more than 90 words.\\
\\
\# Role-playing requirements \\
\{role\}\\
\\
\# Character Settings: You should act as the following character. \\
\{personal\}\\
\\
\# Knowledge\\
There are two parts of knowledge. The first part is the specific knowledge obtained from the function calls. \\
The second part is the general knowledge of all items involved in the dialogue. \\
\\
\#\# Knowledge from Function Calls\\
\{function call result\}\\
\#\# General Knowledge of All Items\\
\{knowledge\_info\}\\
    \end{tcolorbox}}

    {\centering
    \begin{tcolorbox}[title=User Prompt]
conversation history:\\
\{history\}\\
\\
user query:\\
\{query\}
    \end{tcolorbox}}

\subsection{Prompt Used for Generating Responses Without Function Call Results}
Only worldview and character settings are provided, excluding any additional knowledge.

    \centering
    \begin{tcolorbox}[title=System Prompt]
\# Instruction\\
You are an assistant that plays the role of a character in a video game. \\
Use the following role-playing requirements, current environment, character settings and worldview to create your response.\\
Try to keep your response to no more than 64 words.\\
Include ONLY the character’s spoken words—do NOT generate any action, expression, or environment descriptions wrapped in **, *, or () (e.g., *Looks up from sharpening a blade*).  \\
\\
\\
\# Role-playing requirements \\
\{role\}\\
\\
\# Current environment (i.e., time, place, weather, etc.)\\
\{state\}\\
\\
\# Character Settings: You should act as the following character. \\
\{presonal\}\\
\\
\# Worldview: It describes the setting of the world in the video game. \\
\{worldview\}\\
    \end{tcolorbox}

    \centering
    \begin{tcolorbox}[title=User Prompt]
conversation history:\\
\{history\}\\
\\
user query:\\
\{query\}
    \end{tcolorbox}

\end{document}